\documentclass[letterpaper, 10 pt, conference]{ieeeconf}  

\IEEEoverridecommandlockouts                              

\usepackage{graphics} 
\usepackage{epsfig} 
\usepackage{mathptmx} 
\usepackage{times} 
\usepackage{amsmath} 
\usepackage{amssymb}  
\usepackage{hyperref}  
\usepackage[T1]{fontenc}

\title{\LARGE \bf Predictive and adaptive maps for long-term visual navigation\\ in changing environments}

\author{Lucie Halodov{\' a}$^1$, Eli{\v s}ka Dvo{\v r}{\' a}kov{\' a}$^1$, Filip Majer$^1$, Tom{\'a}{\v s} Vintr$^1$,\\ Oscar  Martinez Mozos$^2$, Feras Dayoub$^3$, Tom{\'a}{\v s} Krajn{\'i}k$^1$
	\thanks{$^{1}$Artificial Intelligence Center, Czech Technical University {\tt\small{{halodluc,krajnt1}@fel.cvut.cz}}}
	\thanks{$^{2}$Technical University of Cartagena, Spain}
	\thanks{$^{3}$Australian Centre for Robotic Vision, QUT}
	\thanks{The work was funded by the Czech Science Foundation project 17-27006Y and by Spanish Ramon y Cajal Programme (ref. RYC-2014-15029)}
}

\begin{document}
\maketitle

   \begin{abstract}
In this paper, we compare different map management techniques for long-term visual navigation in changing environments.
In this scenario, the navigation system needs to continuously update and refine its feature map in order to adapt to the environment appearance change.
To achieve reliable long-term navigation, the map management techniques have to (i) select features useful for the current navigation task, (ii) remove features that are obsolete, (iii) and add new features from the current camera view to the map.
We propose several map management strategies and evaluate their performance with regard to the robot localisation accuracy in long-term teach-and-repeat navigation.
Our experiments, performed over three months, indicate that strategies which model cyclic changes of the environment appearance and predict which features are going to be visible at a particular time and location, outperform strategies which do not explicitly model the temporal evolution of the changes. 

\end{abstract}
		
   \section{Introduction}

For mobile robots, the most important capability is navigation, i.e. the ability to get to a desired destination.
To navigate in an efficient, reliable and safe way, a mobile robot must be able to determine its position relatively to the path it needs to traverse in order to reach the intended goal location.
To be able to determine its location, a robot has to gather information about its surroundings through sensors.
This information is typically integrated in an environment model, which is, in turn, used to calculate the robot position once the robot acquires new sensory measurements.

One of the main challenges faced by localisation and map building systems is the noise of the information acquired by the robot. One source of noise is the uncertain and incomplete data provided by its sensors. This problem, though, can be overcome by methods that model the uncertainty of the sensor and allow to build maps of large environments and to precisely localise mobile robots different environments~\cite{thrun}.

\begin{figure}[t!]
\begin{center}
\includegraphics[width=0.23\textwidth]{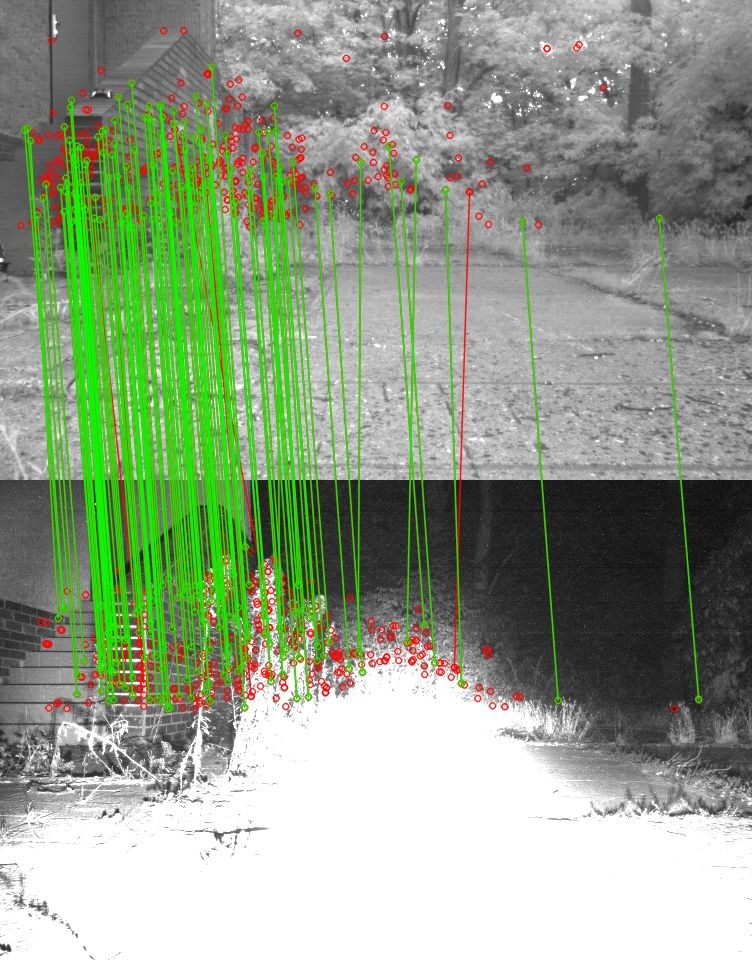}
\hfill
\includegraphics[width=0.23\textwidth]{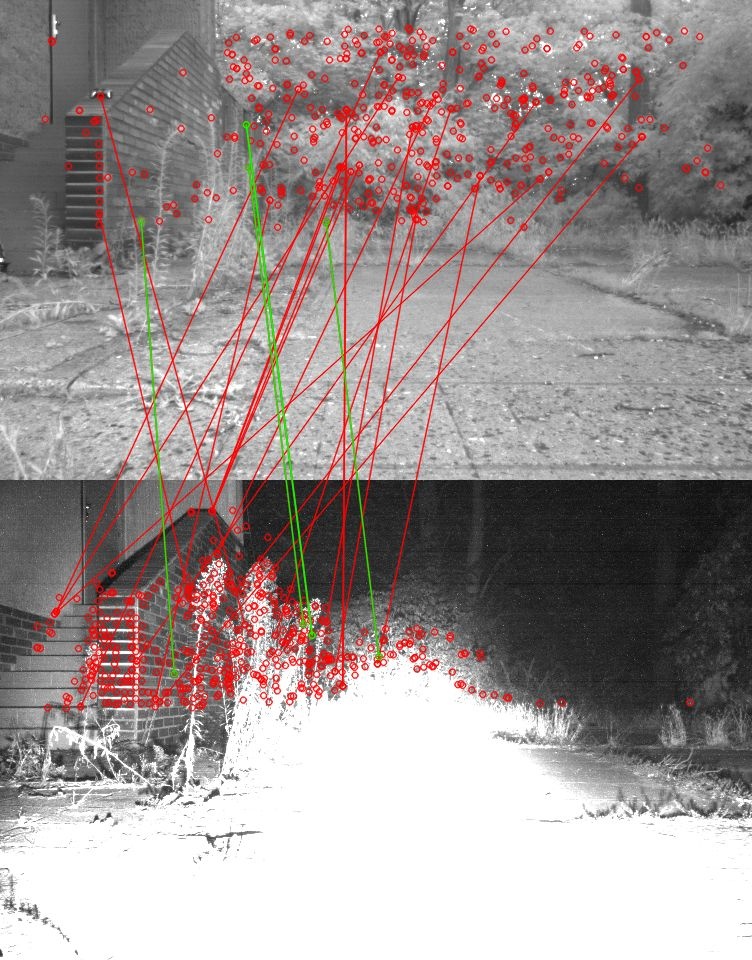}
\caption{Feature matching between the map and the current robot view: The first image shows that the use of score-based map adaptation strategy results in more correspondences compared to the use of a static map, shown in the second image. The green lines represent correct matches, while red lines incorrect ones. 
The score-based map adaptation ensures a sufficient number correct correspondences due to using information from the static map augmented with features from later traversals. \label{pic:mapView}} 
\end{center}
\end{figure}

However, as soon as the robots should navigate for longer periods of time, another source of uncertainty appeared. This uncertainty is caused by the fact that all environments change over time, either through natural processes or as a result of human activity. Thus, the localisation and mapping system of a mobile robot that has to autonomously operate for longer periods of time has to deal with two inherently different sources of uncertainty: the one caused by the environmental changes, and the one caused by the sensor noise.

Accumulation of either uncertainty in the map will negatively affect localisation accuracy and navigation reliability, which are crucial for efficient mobile robot operation.
On the one hand, updating the entire map with new information is prone to sensor noise.
As shown in~\cite{ulmann}, localisation errors caused by the sensors noise slowly, but gradually accumulate and decrease the map accuracy up to the point where the localisation fails to achieve the required accuracy. 
On the other hand, a map which is not updated with new information becomes gradually irrelevant due to the environment changes.
Thus, to achieve long-term operation, a mobile robot has to update and refine the environment map in a way that suppresses the effect of sensor noise, while also dealing with the effects of the changes in the environment.
First map update has to include new features from the camera view to ensure adaptation to changes.
Second, the map should be refined so that only the most relevant features are used for localisation. 
Finally, obsolete map features which slow down the feature matching and might cause perceptual aliasing, have to be removed from the map.

In this paper we test different strategies to update the map during long term navigation.
We based our comparison on the teach-and-repeat paradigm, where a robot is able to repeat a path which was previously taught by a human operator. This paradigm has reported to achieve reliable visual navigation in outdoor environments for longer time periods~\cite{bearnav,jfr10,mesasb}.

This paper presents several contributions related to teach-and-repeat navigation systems operating in naturally changing environments. A first theoretical contribution answers three questions related to visual feature management:
\begin{itemize}
\item Which features should be discarded from the map?
\item How to select the most relevant features?  
\item Which features should be added to the map?
\end{itemize}

In other words, the paper provides insights about which features to remember, which to use at a given situation, and which to forget.
To answer these questions, we provide a basic formalisation of the problem, suggest performance metrics to evaluate various feature update strategies, and experimentally evaluate them on datasets gathered by a real mobile robot in outdoor environments over the period of three months. 

A technical contribution of the paper is a simple-to-use, reliable navigation system, which is able to continuously adapt maps to keep up with the environment changes, providing a robot with the ability of continuous operation.
The system is available at~\cite{system} as an open-source ROS package along with the datasets and codes that allow to reproduce the experiments presented here.

   \section{Related work}

Up to our knowledge, one of the earliest papers on long-term mobile robot operation was addressed in~\cite{awesome}.
Here, the authors already mention that the robots will have to be able to deal with the fact, that most environments are not static, but are subject to perpetual changes.
Later papers, such as~\cite{valgren} investigated the long-term stability of image features in outdoor environments and \cite{monocular,jfr10} investigated the impact of seasonal changes on the performance of teach-and-repeat systems.
The \cite{monocular} concluded, that to cope with seasonal changes, teach-and-repeat systems would benefit from the use of multiple maps, which would be selected by means of mutual information entropy~\cite{hanka}.

One of the most popular frameworks that uses multiple maps is the experience-based localisation~\cite{churchill}, which represents the same location with several appearances or `experiences'.
These depend on particular environmental conditions such as day/night or weather.
In this framework, a robot retrieves several `experiences' tied to a given location and tries to associate them with its current view. 
A failed association indicates that the appearance of the particular location changed drastically and the perceived appearance is added to the set of experiences associated with a given location. 
The experience-based approach allows smart management of the environment maps~\cite{gadd}, selecting experiences \cite{smart} is similar to our work by discarding fatures from the map.
\cite{paton} describes successful integration into teach-and-repeat systems. 

Another framework based on the 'experiences'~\cite{churchill} is described in \cite{paton2016bridging}, in which multiple-experience localisation is performed, but the `privileged' experience created during the teaching phase is prioritised to prevent accumulation of errors in the map by long-term drift (\cite{paton} refers to the problem as photocopy of a photocopy).
New experiences are transformed to the frame of `privileged' one along with their uncertainties, to support the localisation of the robot relatively to the `privileged path'. 
\cite{mactavish2018selective} summarizes previous experience-based techniques and introduce a novel collaborative filtering method, capable of recommending the most relevant experiences, which improves their system performance.

Another popular approach, inspired by the interplay of long- and short-term memory, is described in \cite{dayoub2008,biber}, who gradually add newly detected features and discard features which are no longer visible. 
In \cite{dayoub2008}, the system of feature exchange is based on a state automaton, in which all newly detected features are passed into a short-term memory.
Those that are repeatedly detected in short-term memory pass through several states, finally ending in long-term memory. 
Features becoming gradually obsolete return back through these states until they are removed from the map, i.e. they are completely forgotten. 
On contrary, in \cite{biber}, short-term memory is updated online each time new sensory data are obtained and long-term memory is updated offline, typically once a day. 
Their system is based on randomly selecting sample values from sensory data and probabilistic and statistical methods to update long-term memory. 

A similar approach group based on ranking features is described in \cite{summary_maps,burki2018map}, who to remove or add the elements to the current map based on their ranking. 
In \cite{summary_maps}, a framework called Summary Map is presented, in which all new features are added to the database, from which a smaller subset is selected based on the ranking function. 
The function is based on past visibility of map features and the size of database is limited by removing features with large reprojection errors. 
The concept of database is also used in \cite{burki2018map}, in which summarization over the database is performed to ensure fixed map size. 
The ranking function based on the probability of feature visibility in current conditions and selects features as additional information to be able to distinct usefulness of landmarks. 
If a localisation is not successful, summarization of new set of features is performed as an optimisation problem.

Once the navigation can cope with the changes, the robot can gather enough data to learn about long-term temporal behaviour of the environment.
Lowry et al. \cite{lowry2016supervised} uses consecutive observations identify time variable and time-invariant image components, Rosen et al. \cite{rosen2016towards} use long-term observations to predict, which image elements are more likely to disappear after some time.
Krajn{\'\i}k et al. \cite{krajnik2017fremen} propose to use spectral analysis to capture the periodic behaviour of the changes caused by day/night and seasonal processes.
Similarly \cite{neubertappearance,cnn_horia,cnn_latif} attempt to predict how a summer scene will look like based on its appearance captured during winter and vice versa.

Since the problem of long-term, reliable operation gradually comes into focus of the robotics research community, the aforementioned list of papers is by no means exhaustive.
Thus, we refer to a comprehensive review of approaches for long-term visual localisation in \cite{lowry2016visual} and a general review of AI methods for long-term autonomy in~\cite{lta2018review}.

Inspired by~\cite{dayoub2008,biber,summary_maps,krajnik2017fremen}, we implemented similar schemes, which are capable to gradually adapt the map during the routine operation of the robot.

   \section{Visual teach-and-repeat navigation}\label{sec:navsys}

Our navigation system is based on the teach-and-repeat paradigm. 
This paradigm is popular in industrial robotics. Here an operator teaches a robot to perform some task by guiding its arm along the desired path. Afterwards the robot repeats the same movement autonomously.

The teach-and-repeat navigation for mobile robots have shown to perform well in long-term scenarios~\cite{barfoot,bearnav,surfnav,paton} and in difficult illumination conditions~\cite{paton2016dead}.

In our implementation~\cite{bearnav}, during the teaching phase, an operator guides a mobile robot which creates a series of local maps at regular distances, that are measured by odometry. Each local map contains a captured image with a set of visual features extracted by the SURF detector~\cite{surf} and the BRIEF descriptor~\cite{BRIEF}, which have shown their suitability for long-term teach-and-repeat visual navigation~\cite{grief}. As a result, the path followed by the mobile robot is represented by a sequence of local maps indexed by their odometry distance from the starting point.
%
 
During the repeat phase, the robot retrieves the local map according to its odometry distance from the starting point and matches the maps features to the new ones extracted from its current camera image.
After that, it subtracts the horizontal image coordinates of the established feature pairs.
Then it employs a simple histogram voting scheme to calculate the most prevalent difference $\delta$.
This process efficiently recovers the horizontal shift $\delta$ between the currently seen scene and the local map image, i.e. it is equivalent to image registration in the horizontal axis.
Since the features detected in the image do not lie in infinity, the value of $\delta$ indicates not only the difference between the current heading and previous taught heading, but also the robot lateral displacement from the path.
Therefore, if the robot steers to maintain $\delta$ low, then it keeps itself close to the taught path.
The works~\cite{bearnav,jfr10,mesasb} provide a mathematical proof and experimental evidence showing that the aforementioned navigation scheme, while being rather simple, efficiently suppresses both lateral and longitudinal errors of the robot position and keeps the robot on the taught path. Further details, codes and videos of the navigation system are also provided at \cite{system}.

In this implementation, during the repeat phase, the robot does not update the features in its map and it always uses as reference the initial map created during the teaching phase. While this strategy showed good performance in several environmental conditions, it was not able to handle significant appearance changes, e.g. day/night or winter/summer. Therefore in this paper we extended the repeat phase with different map adaptation strategies.

   \section{Map management}\label{sec:methods}

To achieve long-term operation, we extended the aforementioned system by the ability to adapt its maps during the repeat phase.
The core idea is to evaluate the utility of the local map features, keep the useful ones, remove the features which are often matched incorrectly, and add features which were not visible before.

To do so, we monitor the contribution of the features stored in the local maps to the quality of the navigation.
As mentioned in Section~\ref{sec:navsys}, our navigation system continuously retrieves the image features from the local maps in the robot vicinity, matches these features to the ones extracted from the camera image, and uses a histogram voting method to determine the most frequent difference in the horizontal coordinate of the feature pairs.
When the aforementioned system establishes correspondences between the local map features and the currently-visible ones, each local map feature is assigned one of the three states:

\subsubsection{Not matched feature}
In most cases, the map feature is \textit{not matched} to any visible one, i.e. there is no correspondence.

\subsubsection{Correctly matched features}
Matched pair is assumed to be~\textit{matched correctly} if their difference in horizontal coordinate is consistent with the results of histogram voting, i.e. it belongs to the most frequent difference.
These features are used to calculate the shift $\delta$ between the local map image and the currently visible one.

\subsubsection{Incorrectly matched features}
If the difference in horizontal coordinate is not consistent with the results of the histogram voting, i.e. it does not belong to the most frequent differences, then these features forming a pair are assumed to be~\textit{matched incorrectly}.

An ideal map update scheme would keep the features which are more likely to be \textit{matched correctly} and
 substitute the ones which are typically~\textit{matched incorrectly} or \textit{not matched} with features which were not detected before.
However, the aforementioned states only refer to the past usefulness of the particular image feature and they can't precisely predict how useful the feature will be in the subsequent runs.
In other words, while the works presented in~\cite{summary_maps,krajnik2017fremen} indicate that if a feature was well visible and paired in the past, it is more likely to be visible in the future, one cannot take that for granted.

\subsection{Discarding map features}

The first question we investigate in this paper is how to use the information about past feature performance to determine if we should keep it in the map and how influential the feature should be on the localisation process.
Thus, we implemented several map update schemes, and evaluated them in real-world experiments performed over the period of three months.
Each update scheme exploits the information about the past feature performance in a different way.

\subsubsection{Static map strategy}

The original method, used as a baseline in our experiments, simply ignores the results of the feature matching history and does not update the map build during the teaching phase.
This corresponds to the system originally proposed in~\cite{bearnav,jfr10}, which could not properly handle significant changes in environment appearance.
Thus, we hypothesise that this map will achieve good accuracy in scenarios with low appearance changes, but will fail in longer term.

\subsubsection{Latest map strategy}

The other extreme handling of the newly-perceived information is to discard all features in the previous map and create a new map, which contains the features detected in the currently acquired image.
The problem of the latest map strategy is that since the robot heading is never estimated with an absolute accuracy, the positions of the new features will be slightly offset relatively to the original map.
As this will happen every time the robot repeats the path, the error will accumulate.
A theoretical analysis of the error accumulation problem is presented in~\cite{ulmann} and the field experiments described in~\cite{paton}, refer to this as a `photocopy of a photocopy' problem.
Thus, while the method will be robust to environment changes, the accuracy of the navigation will gradually deteriorate.
Since the aforementioned error can be very subtle, the problem might emerge only after a substantial number of autonomous traversals.
Moreover, as pointed out by~\cite{ulmann,mactavish2018selective} the error accumulation might not even exhibit itself in cases where the robot traverses multiple times a straight trajectory.
Furthermore, if the localisation will fail (e.g. because of a dynamic object obstructing most of the robot view), the features added in the new local map will be on incorrect positions, breaking the local map.
In other words, this map update scheme is sensitive to sensor noise, occasional localisation failures etc..
No matter how insignificant or rare these are, the long-term drift will cause their accumulation, gradually decreasing the map quality.

\subsubsection{Aggressive map strategy}

The aggressive update scheme discards all features which were \textit{not matched} or \textit{matched incorrectly} and substitutes them with features detected in the currently acquired image.
We hypothesise that similarly to the previous case, the accuracy of this map update scheme will suffer from a long-term drift caused by localisation inaccuracy.
Furthermore, this map update is also prone to localisation failures.

\subsubsection{Strict map strategy}

The strict update scheme discards all features which were \textit{matched incorrectly} and substitutes them with the same number of features detected in the currently acquired image.
We hypothesise that unlike in the previous case, the drift will be suppressed, but the adaptation to the change might be rather conservative and the map might have trouble adapting to the environment change.

\subsubsection{Summary map strategy}

The summary update scheme discards all features which were \textit{matched incorrectly} and adds 10\% of the features detected in the currently acquired image.
The map will grow in size over time, which gradually deteriorates the real-time response of the feature matching and might result in perceptual aliasing.

\subsubsection{Multiple map strategy}

This strategy is inspired by the methods, which maintain multiple representations of the same location and determine the position using a subset of those, such as~\cite{churchill,smart}.
In particular, when the number of features that are \textit{matched correctly} falls below a certain threshold, we store the currently detected features in an alternative local map.
During autonomous navigation, the current view is matched to all of the maps associated with the same location and the map with the most matches is used to determine the robot steering.
If none of the maps provide enough features which are \textit{matched correctly}, another local map is added again.

\subsubsection{Score-based, adaptive map strategies}

This `adaptive' map update scheme is inspired by the works~\cite{dayoub2008,biber,summary_maps}, which remove or add features based on their past influence on the localisation quality.
In this map update scheme, each feature is assigned a score, which is increased by $s_c$ when the feature is \textit{matched correctly} and decreased by $s_i$ when the feature is \textit{matched incorrectly} and by $s_n$ when the feature is \textit{not matched}.
Each time a robot used a given local map for navigation, it updates the score of the features, removes the $n$ features with the worst score and substitutes them with the features from the current view.
The crucial questions here are how to select the number of features to be exchanged (i.e. $n$) and the scores $s_c$, $s_i$ and $s_n$.
One of the important questions is the choice of $n$, which defines how quickly the map adapts to the changes.
If the $n$ is low, the map cannot adapt to fast environmental changes, but it's robust to occasional glitches of the image registration, which might result in wrong features being added to the map.
Choosing a high $n$ makes the map able to adapt to rapid environment changes, but it may result in gradual deterioration of the map.
Moreover, a failure of the image registration populates the map with $n$ features at wrong, but consistent positions, which might corrupt the local map.
Ideally, $n$ should be high enough to ensure that the map can keep up with the environment variations, but it should be small enough ensure that occasional failure of localisation does not corrupt the map.
The choices of $n$, $s_c$, $s_i$ and $s_n$ are investigated in experimental evaluation.

\subsubsection{Frequency Map Enhancement (FreMEn) strategy}
\begin{figure*}[!ht]
\begin{center}
\includegraphics[width=0.95\textwidth]{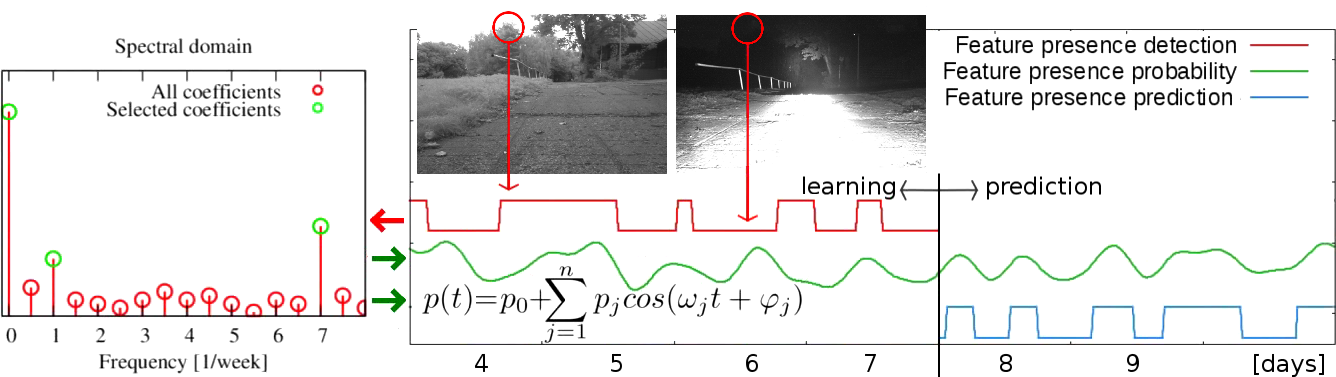}
	  \caption{Frequency Map Enhancement (FreMEn) for visual localisation: The observations of image feature visibility (centre,red) are transferred to the spectral domain (left). The most prominent components of the model (left,green) constitute an analytic expression (centre,bottom) that represents the probability of the feature being visible at a given time (green). This is used to predict the feature visibility at a time when the robot performs self-localisation (blue). For further details, see a video at~\url{https://youtu.be/Qw1kS_5zVwE}, repository at \url{fremen.uk} or article \cite{krajnik2017fremen}.}\label{fig:fremen}
\end{center}
\end{figure*}

As pointed out in~\cite{rosen2016towards,krajnik2017fremen}, considering the periodicity and persistence of the feature visibility results in a significant improvement of the localisation robustness.
Thus, we decided to apply the Frequency Map Enhancement (FreMEn)~\cite{krajnik2017fremen}, which extracts patterns of the environment changes and uses them to predict the future environment states, see the idea illustrated in Fig.~\ref{fig:fremen}.
The method approximates temporal evolution of a given state by a combination of harmonic functions, which are identified by means of spectral-frequency analysis.
Thus, we extended the definition of the scores from the previous case to the time domain, which means that instead of just increasing or decreasing the feature score $s_c$ by $s_i$, $s_n$ and $s_c$ based on the last matching results, we also take into account the time of the matching $t_k$.
Thus, each time a feature is \textit{matched correctly}, \textit{matched incorrectly} or \textit{not matched}, we add the corresponding $s_c(t_k)$, $s_i(t_k)$ or $s_n(t_k)$ to the FreMEn model of that feature.
When the robot starts to perform autonomous navigation, we use the current time $t$ to predict the score of each feature and select $m$ of the highest-scoring features to be used for localisation.
The time dependent score reflect the fact, that the visibility of each feature is strongly influenced by the time of the day, season of the year etc.
Thus, the FreMEn strategy will select different features based on the time the robot needs to navigate autonomously.
To identify obsolete features, we calculate the mean score of each feature across time, remove $n$ features with the lowest score and substitute them with new ones.

\subsection{Adding new scene features}

Another question which we have to answer is which of the previously-unseen features to add to the map.
In our case, we score the unmatched `view' features, i.e. features extracted from the current camera image, according to the distance of their descriptors to the features of the map.
In particular, the score of each view feature equals to its distance (in descriptor space) to the nearest feature in the local map, which is related to the feature's uniqueness.
Highly-scored view features, which are unlikely to be mismatched and paired incorrectly, are good candidates to be added to the map.
Thus, when choosing the features to add to the map, one order the unmatched features

In our experiments, selecting the features to add by their descriptor distance provided statistically significantly better results than a random selection across all the tested strategies.
Thus, in all the following strategies, the features to be added are selected based on the descriptor-distance-based score.
When the features are added to the map, their positions in the current view are first corrected by the current horizontal shift $\delta$, calculated by the histogram voting, see Section~\ref{sec:navsys}.

   \section{Experimental evaluation}

To evaluate how the aforementioned map adaptation methods affect the performance of the visual navigation in changing environments, we performed two experiments with a CAMELEON tracked robot.
The robot was equipped with an aluminium superstructure with several mounts for the TARA camera, intel i3 control laptop, 4000 lumen searchlight and an A3-sized pattern for precise measurement of the robot position by an external localisation system~\cite{whycon}.
The experimental evaluations were performed in a small forest park around the Hostibejk site in Kralupy nad Vltavou, Czechia, see~\cite{system}.
The evaluation is based on a dataset which was gathered at the aforementioned location over the course of three months.


To gather the datasets, we taught the robot a short, closed path and during three months, we let the robot to navigate autonomously using the \textit{static}, \textit{latest} and \textit{score-based map} methods in almost 200 sessions.
During the data collections, the robot was supervised or tracked by~\cite{system}, and whenever the navigation failed or diverged too far away, the map update strategy was switched and the data-gathering session was restarted.
Thus, we could qualitatively compare the efficiency and reliability of the aforementioned map update strategies using not only the gathered datasets, but also as in a more realistic scenario, where the map update was used as a component of a vision-in-the-loop system.
While the \textit{latest} strategy was subject to a small drift, which was typically noticeable only after several traversals, the \textit{static} map strategy produced a stable beheviour.
However, as soon as the illumination changed significantly, e.g. from day to night, navigation according to the \textit{static} maps failed.
The \textit{score-based map} was able to guide the robot through data gathering sessions, which started during a late afternoon, proceeded over the evening and ended at night.

\begin{figure}[!ht]
\begin{center}
\includegraphics[width=0.32\columnwidth]{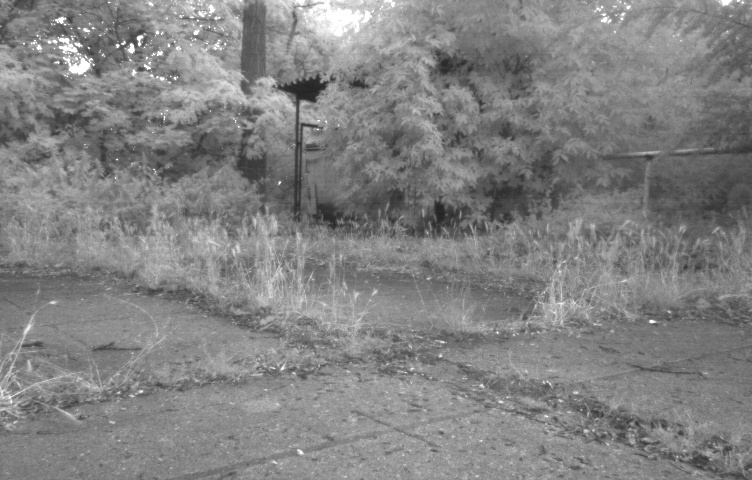}\hfill
\includegraphics[width=0.32\columnwidth]{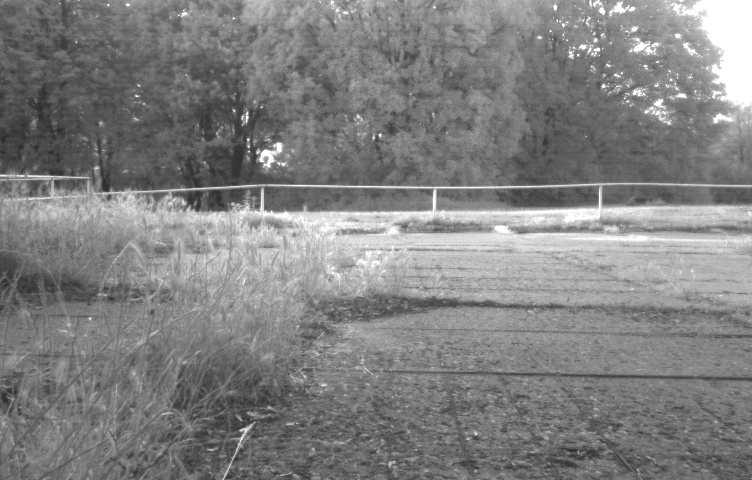}\hfill
\includegraphics[width=0.32\columnwidth]{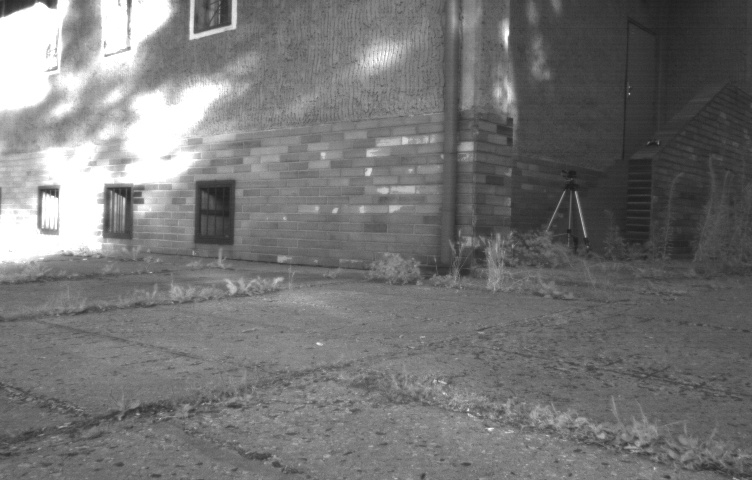}\\
	\vspace{2mm}
\includegraphics[width=0.32\columnwidth]{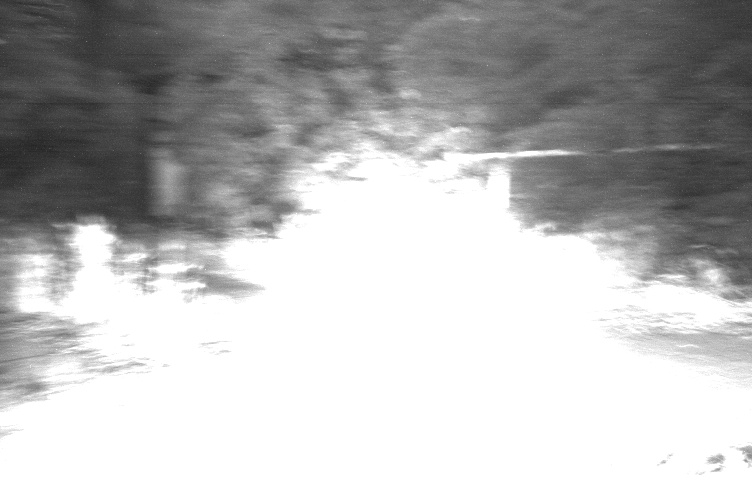}\hfill
\includegraphics[width=0.32\columnwidth]{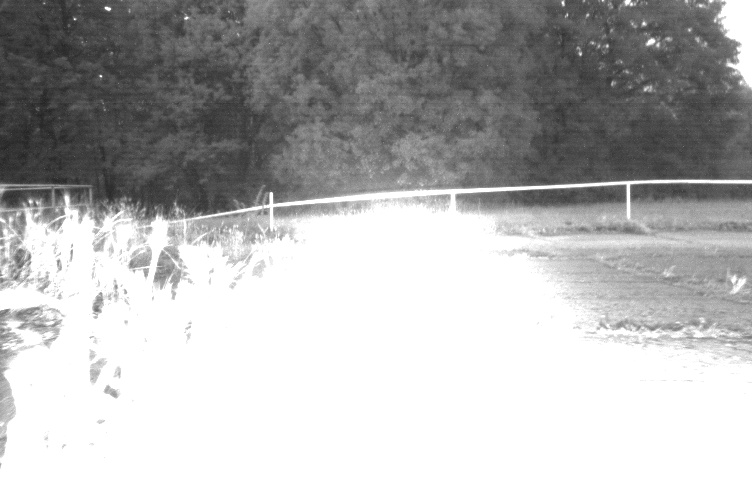}\hfill
\includegraphics[width=0.32\columnwidth]{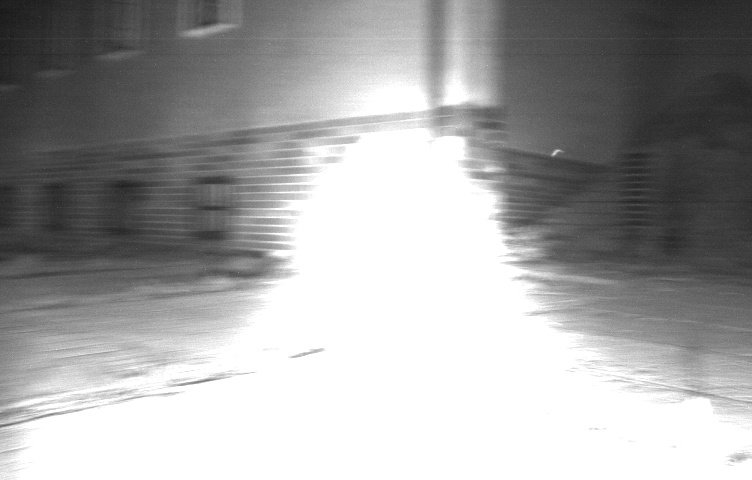}
\caption{Representative images of the dataset collected. Columns correspond to the same locations, with top images obtained during the day and bottom images captured during night.\label{fig:changes}}
\end{center}
\end{figure}

Apart from day/night transitions, the dataset also contains diverse environmental conditions ranging from cloudless days to light rains.
During these autonomous traversals, the robot not only updated its map, but it also saved the images from which the new features were obtained.
Most of the images contain trees, shrubs and other natural structures, and approximately one-third of the local maps also captured parts of a small building at the Hostibejk site.
Since the taught path was represented by 32 local maps, and the robot traversed the path 178 times, the resulting dataset is composed of 5696 images.
The dataset is available at~\cite{system} and the appearance changes are documented in Figure~\ref{fig:changes}.

\subsubsection{Evaluation methodology}

The primary output of our navigation system is the steering correction, which is determined by registering the images captured by the robot camera with images from the local maps.
Thus, the precision of image registration directly influences accuracy and reliability of visual-based navigation.
Therefore, evaluation of the various map update schemes can be based on the accuracy of the image registration.

In particular, we extract the images captured during the autonomous drive from the dataset and simulate the robot movement by `replaying' them to the robot navigation system, which processes them in the same way as during normal operation.
Then, we compare the calculated shift $\delta$ to the ground truth, obtained by manual annotation using a specialised tool for image registration~\cite{system}.
A previous work, presented in~\cite{hanka} indicated that the results of the manual annotation are consistent across several users of the aforementioned tool.
The difference between the shift calculated by the system and the human-generated shift is used as a measure of the method accuracy.
Since during each simulated drive, the system registers several hundreds of images, we denote the result of $i^{th}$ image registration as $\delta_i$ (see Section~\ref{sec:navsys}).
By comparing the calculated $\delta_i$ to the values obtained by human annotation $\gamma_i$, we obtain a sequence $\epsilon_i = |\delta_i - \gamma_i|$ which corresponds to the accuracy of the image registration for $i^{th}$ image pair.

Thus, evaluation of $j^{th}$ method or its particular setting produces an error sequence of $\epsilon_i^j$.
To determine which method achieves more accurate navigation, we compare the values $\epsilon_i^j$ calculated by the aforementioned procedure qualitatively and statistically.
For statistical evaluation, we apply Student's paired sample test, which is able to determine if an error sequence of $\epsilon_i^A$ generated by method $A$ is statistically significantly smaller than another error sequence $\epsilon_i^B$, generated by method $B$.
For qualitative evaluation, we use the values of $\epsilon_i$ to calculate a function $p(\epsilon_t)$, which indicates the probability that the registration error was lower than a given threshold, i.e. $p(\epsilon) = P(\epsilon_i \leq \epsilon_t)$.
While the quantitative evaluation clearly shows if one method performs better than the other, displaying $p(\epsilon_t)$ of two different methods inside of the same graph provides one with a deeper insight how much these methods actually differ.

\subsubsection{Evaluation results}

\begin{figure}[!ht]
\begin{center}
\includegraphics[width=0.95\columnwidth]{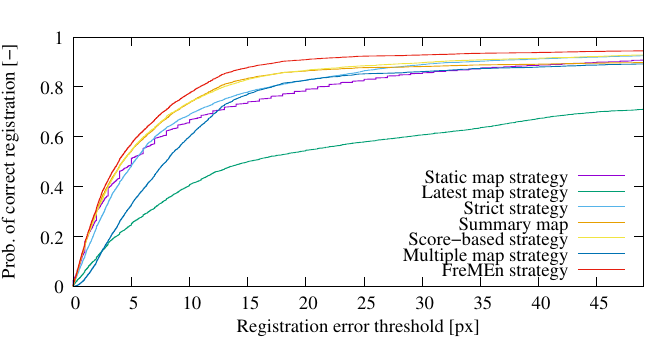}
\caption{Probability of registration error being smaller than a given number of pixels for different map update strategies.\label{fig:experiments}}
\end{center}
\end{figure}

To evaluate the map update schemes, we replayed the dataset images through the teach-and-repeat navigation system, which was using the map update schemes described in the Section~\ref{sec:methods}.
For the \textit{score-based} and \textit{FreMEn} strategies, we used different settings of the features to be exchanged $n$, and feature penalisation $s_c$, $s_n$, $s_i$.
First, we tested if one should penalise features which were not matched several times by setting the value of $s_n$ (penalisation of \textit{not matched} features) to a fraction of $s_i$ (penalisation of features \textit{matched incorrectly}.
Second, we tested several different schemes of setting $n$, which determines how many features are going to be discarded from the map and substituted by the newly-perceived features.
In total, we run 64 simulated runs on the dataset's images, obtaining localisation errors of all the strategies mentioned in Section~\ref{sec:methods} and several different settings of the \textit{adaptive map}.
Using the Student's paired t-test, we identified strategies which performed statistically significantly better than other ones.
The best performing map update strategy was the \textit{FreMEn} one, followed by the \textit{score-based} one.
Both of these strategies explicitly penalised only the features that were \textit{matched incorrectly} and the score of \textit{not matched} features set to 0, i.e. its $s_c(t) = s_i(t) = 1$ and $s_n(t) = 0$.
Both methods were set to provide 500 features per local map to the navigation system and exchange 5\% of features per run.
To have a deeper insight into the performance of the individual methods, see Fig.~\ref{fig:experiments}, which shows the cumulative probabilistic distribution of registration error for each of the strategies investigated.
To make this experiment repeatable, we provide codes, datasets and additional information at~\cite{system}.
One of the surprising results of the experiments was the rather low performance of the \textit{latest map}.
We assume that the observed drift was caused by combination of several factors: adverse lighting conditions, causing significant image blur at areas there the robot had to turn rapidly, real-time issues of the low-performance computer, rather inexpensive camera etc.
Moreover, the effects of the drift started to be prominent only after a high number of traversals.

   \section{Conclusion}

In this paper, we present a comparison of map adaptation strategies for mobile robots that have to operate for extended time periods in outdoor natural environments.
To perform the comparison, we integrated several map adaptation strategies into a teach-and-repeat visual navigation system and used it to repeatedly guide a mobile robot along the same path for the period of three months.
During these traversals, the robot had to deal with weather ranging from cloudless days to light rain, and illumination conditions ranging from bright days to full nights, where the robot had to use its own source of illumination.
The data gathered were used for thorough comparison of the various map adaptation strategies and their various settings.
The experimental evaluations indicate that blind use of the most recent information leads to quick deterioration of map quality and navigation accuracy.
The best performing strategies were based on temporal analysis~\cite{krajnik2017fremen} and gradual map adaptation similar to the methods presented by~\cite{dayoub2008}.
To ensure the reproducibility of the research presented, we provide the system's source codes as well as access to the datasets used for evaluation at \cite{system}.

In future work, we would like to scale up the experimental verification by evaluating the aforementioned strategies using publicly available datasets and a high-fidelity simulators.
In particular, we plan to evaluate the results using the Oxford's Robotic Car and EU Long-term datasets~\cite{oxfordDataset,utbm_robocar_dataset}.
These consist of several repetitions of the same path in Oxford and Paris including various weather conditions, and both short- and long-term changes.
To allow even more path repetitions,  we will use the CARLA Simulator~\cite{carlaPaper}, which can be seamlessly integrated with our system, because of its compatibility with ROS.

\bibliographystyle{IEEEtran}
\bibliography{main}

\end{document}